\documentclass[11pt,a4paper]{article}

\usepackage[utf8]{inputenc}
\usepackage[T1]{fontenc}
\usepackage{amsmath,amssymb}
\usepackage{graphicx}
\usepackage{booktabs}
\usepackage{hyperref}
\usepackage[margin=1in]{geometry}
\usepackage{xcolor}
\usepackage{enumitem}
\usepackage{listings}
\usepackage{natbib}
\usepackage{caption}
\usepackage{subcaption}
\usepackage{float}
\usepackage{array}
\usepackage{algorithm}
\usepackage{algpseudocode}
\usepackage{tikz}
\usetikzlibrary{shapes.geometric,arrows.meta,positioning,fit,calc}

\lstdefinelanguage{yaml}{
  keywords={true,false,null,y,n},
  keywordstyle=\color{blue!60!black}\bfseries,
  sensitive=false,
  comment=[l]{\#},
  morecomment=[s]{/*}{*/},
  commentstyle=\color{gray!70!black}\ttfamily,
  stringstyle=\color{green!40!black}\ttfamily,
  moredelim=[l][\color{magenta!60!black}]{\&},
  moredelim=[l][\color{red!60!black}]{*},
  morestring=[b]",
  morestring=[b]'
}

\newcolumntype{L}[1]{>{\raggedright\arraybackslash}p{#1}}

\hypersetup{
  colorlinks=true,
  linkcolor=blue!70!black,
  citecolor=blue!70!black,
  urlcolor=blue!70!black
}

\title{\textbf{Demand-Driven Context: A Methodology for Building Enterprise Knowledge Bases Through Agent Failure}}

\author{
  Raj Navakoti \\
  \texttt{rajnavakoti@gmail.com}
  \and
  Saideep Navakoti \\
  \texttt{Saideep.dev@outlook.com}
}

\date{March 2026}

\begin{document}

\maketitle

\begin{abstract}
Large language model agents demonstrate expert-level reasoning and processing capabilities, yet consistently fail on enterprise-specific tasks due to missing domain knowledge---terminology, operational procedures, system interdependencies, data structures, and institutional decisions that exist largely as tribal knowledge. Current approaches fall into two categories: top-down knowledge engineering, which attempts to comprehensively document domain knowledge before agents use it, and bottom-up automation, where agents learn from their own task experience to improve execution strategies. Both have fundamental limitations: top-down efforts struggle with discovery---in large enterprises, knowing what knowledge exists and where it resides is itself a hard problem---and produce bloated, untested knowledge bases; bottom-up approaches optimize how agents behave but cannot acquire domain knowledge that exists only in human heads. We present \textbf{Demand-Driven Context (DDC)}, a problem-first methodology that uses agent failure on real problems as the primary signal for what domain knowledge to curate. Inspired by Test-Driven Development, DDC inverts the knowledge engineering process: instead of curating knowledge and hoping it is useful, DDC gives agents real problems, lets them demand the context they need to solve those problems, and curates only the minimum knowledge required to succeed. We describe the methodology, its entity meta-model, and a convergence hypothesis suggesting that 20--30 problem cycles produce a knowledge base sufficient for a given domain role. We demonstrate DDC through a worked example in retail e-commerce order fulfillment, where nine DDC cycles targeting an SRE incident management agent produce a reusable knowledge base of 46 entities. Finally, we propose a scaling architecture for enterprise adoption that introduces semi-automated curation with human governance, enabling parallel problem-solving across teams.
\end{abstract}

\noindent\textbf{Keywords:} context engineering, knowledge acquisition, LLM agents, enterprise AI, demand-driven curation, domain knowledge management

\vspace{1em}

\section{Introduction}
\label{sec:introduction}

Give a large language model a problem in software architecture, medicine, or law, and it performs remarkably well. Give it a problem specific to \emph{your} enterprise---one involving your internal systems, your domain-specific terminology, your operational workflows, and the decisions that shaped them---and it performs like a new employee on their first day. Expert reasoning. Zero institutional memory.

This is not a model capability problem. Modern LLMs demonstrate strong general expertise across domains---engineering, operations, analysis---and powerful processing capabilities in reasoning and synthesis. Where they consistently score zero is on enterprise domain knowledge: the terminology, system interdependencies, operational procedures, business rules, and institutional history that exist largely as tribal knowledge distributed across wikis, chat threads, runbooks, and the heads of experienced staff.

The enterprise AI community has recognized this gap. Retrieval-Augmented Generation (RAG) attempts to bridge it by retrieving relevant documents at query time~\citep{lewis2020rag}. But RAG retrieves---it does not curate. It surfaces existing documents without verifying their relevance, currency, or completeness. When the knowledge exists only as tribal knowledge---undocumented, fragmented, or outdated---RAG has nothing to retrieve.

More recently, agentic approaches have emerged. ACE~\citep{zhang2025ace} treats LLM contexts as evolving ``playbooks'' that improve through automated feedback loops. Reflexion~\citep{shinn2023reflexion} enables agents to learn from their own failures. ExpeL~\citep{zhao2023expel} extracts reusable insights from agent experience. These systems optimize \emph{how agents behave}---execution strategies, tool-use patterns, reasoning heuristics. But they cannot acquire the enterprise domain knowledge that makes those strategies meaningful. An agent that has learned excellent reasoning strategies still cannot answer ``Why does the nightly batch job retry three times before escalating to the on-call team?'' without someone providing that context.

We propose \textbf{Demand-Driven Context (DDC)}, a methodology that treats agent failure as the primary signal for what enterprise knowledge to curate. The core insight is borrowed from Test-Driven Development~\citep{beck2002tdd}: just as TDD writes a failing test before writing code, DDC gives an agent a failing problem before curating context. The agent's failure identifies precisely what domain knowledge is missing. A human expert then curates only the minimum context needed for the agent to succeed. This curated knowledge is structured using a typed entity meta-model, stored as version-controlled markdown files, and accumulated across cycles.

DDC makes three claims:

\begin{enumerate}
  \item \textbf{Agent failure is a reliable signal} for identifying knowledge gaps in enterprise domains.
  \item \textbf{Human-curated, demand-driven knowledge} is more efficient and more accurate than top-down documentation or automated context optimization.
  \item \textbf{Convergence}: after 20--30 DDC cycles for a given domain role, the knowledge base stabilizes---each new problem requires fewer new entities because previous cycles already curated overlapping knowledge.
\end{enumerate}

In this paper, we describe the DDC methodology (Section~\ref{sec:methodology}), demonstrate it through a worked example in retail order fulfillment incident management (Section~\ref{sec:example}), present a scaling architecture for enterprise adoption (Section~\ref{sec:scaling}), and discuss limitations and future work (Sections~\ref{sec:discussion}--\ref{sec:conclusion}).

\section{Background and Related Work}
\label{sec:related}

DDC sits at the intersection of three established research areas: knowledge acquisition, failure-driven learning for LLM agents, and context engineering.

\subsection{Knowledge Acquisition and Engineering}

The challenge of acquiring domain knowledge for intelligent systems has been studied since the 1980s. \citet{pazzani1986failure} demonstrated failure-driven knowledge base refinement, where system failures triggered targeted refinement of diagnostic heuristics. This principle---failure as a signal for what to learn---is foundational to DDC, though DDC modernizes it for the era of LLM agents and structured markdown knowledge bases rather than expert systems and production rules.

More recently, the Agent-in-the-Loop (AITL) framework at Airbnb~\citep{zhao2025aitl} embedded human feedback directly into live customer support operations. AITL includes a ``Missing Knowledge Identification'' step where human agents flag knowledge gaps that feed back into the continuous learning pipeline. This shares DDC's insight that humans must identify certain knowledge gaps, but AITL is specific to customer support dialogue and does not generalize to a reusable methodology for arbitrary enterprise domains.

CUGA~\citep{ibm2025cuga} similarly uses enterprise agent failures to inject targeted knowledge, demonstrating that failure-driven knowledge enrichment improves agent performance in enterprise settings. However, CUGA focuses on computer-using GUI agents and does not propose a systematic curation methodology.

\subsection{Failure-Driven Learning for LLM Agents}

Several recent works enable LLM agents to learn from their own failures. Reflexion~\citep{shinn2023reflexion} introduces a verbal reinforcement loop where agents reflect on task failures and maintain a persistent memory of reflections. ExpeL~\citep{zhao2023expel} extracts reusable insights from agent experiences, building a growing knowledge base of task-solving strategies. Both systems are fully automated---the agent generates, evaluates, and curates its own knowledge.

\begin{figure}[t]
\centering
\begin{tikzpicture}[
  label/.style={font=\footnotesize, align=center},
  dot/.style={circle, fill, inner sep=2pt},
]
\draw[-{Stealth[length=5pt]}, thick] (-3.5,0) -- (3.8,0);
\draw[-{Stealth[length=5pt]}, thick] (0,-2.8) -- (0,3.2);

\node[font=\small\bfseries, anchor=south] at (0,3.3) {Domain Knowledge};
\node[font=\small\bfseries, anchor=north] at (0,-2.9) {Strategy Knowledge};
\node[font=\footnotesize, anchor=east] at (-3.5,0.3) {Automated};
\node[font=\footnotesize, anchor=west] at (3.5,0.3) {Human-Curated};

\fill[green!6] (0,0) rectangle (3.5,3);
\fill[gray!6] (-3.5,0) rectangle (0,-2.6);

\node[font=\tiny, gray, align=center] at (-1.75,1.5) {Automated\\Domain Knowledge\\(gap)};
\node[font=\tiny, gray, align=center] at (1.75,-1.3) {Human-Curated\\Strategy\\(manual tuning)};

\node[dot, blue!70!black] (ace) at (-2.2,-1.4) {};
\node[label, blue!70!black, anchor=north] at (ace.south) {ACE};

\node[dot, blue!70!black] (ref) at (-1.8,-2.0) {};
\node[label, blue!70!black, anchor=north] at (ref.south) {Reflexion};

\node[dot, blue!70!black] (exp) at (-2.8,-1.7) {};
\node[label, blue!70!black, anchor=east] at (exp.west) {ExpeL};

\node[dot, orange!70!black] (aitl) at (1.5,0.8) {};
\node[label, orange!70!black, anchor=west] at (aitl.east) {AITL};

\node[dot, orange!70!black] (cuga) at (-1.0,0.6) {};
\node[label, orange!70!black, anchor=south] at (cuga.north) {CUGA};

\node[dot, gray] (rag) at (0.5,-0.5) {};
\node[label, gray, anchor=west] at (rag.east) {RAG};

\node[dot, red!70!black, inner sep=3.5pt] (ddc) at (2.5,2.2) {};
\node[font=\small\bfseries, red!70!black, anchor=west] at (ddc.east) { DDC};

\draw[dashed, gray, rounded corners=3pt] (-3.2,-2.4) rectangle (-1.2,-0.9);
\node[font=\tiny, gray, anchor=north west] at (-3.2,-2.4) {execution optimization};

\end{tikzpicture}
\caption{DDC's position in the context engineering landscape. Existing approaches cluster in the automated--strategy quadrant (optimizing agent execution). DDC occupies the human-curated--domain knowledge quadrant, addressing the upstream question of what enterprise knowledge should exist.}
\label{fig:positioning}
\end{figure}
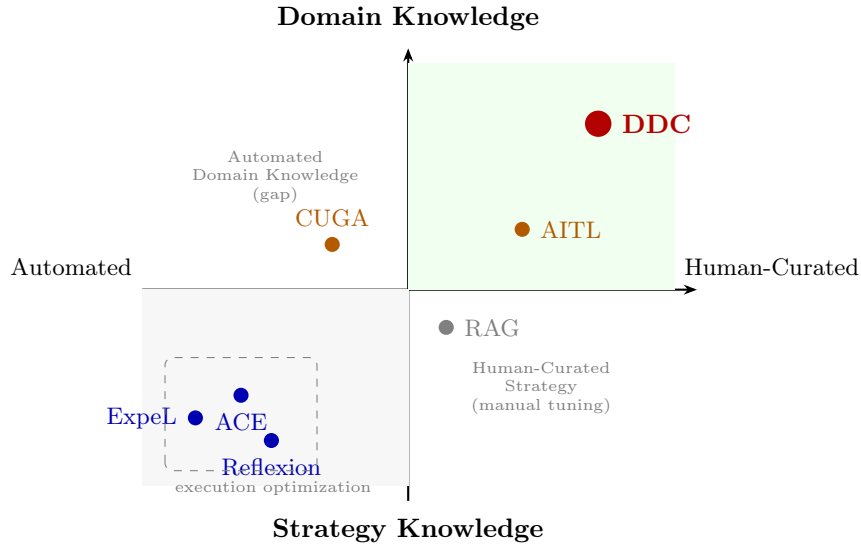

ACE (Agentic Context Engineering)~\citep{zhang2025ace} is the closest related work. ACE treats LLM contexts as evolving ``playbooks'' improved through a three-role architecture: Generator (attempts tasks), Reflector (analyzes failures), and Curator (synthesizes insights into persistent context). ACE achieves +10.6\% on agent benchmarks and +8.6\% on domain-specific finance tasks through fully automated context evolution.

DDC and ACE share the principle of iterative, feedback-driven context improvement. However, they differ in three fundamental ways:

\begin{enumerate}
  \item \textbf{Knowledge type.} ACE optimizes execution strategies---how the agent should behave (tool-use patterns, reasoning heuristics). DDC curates domain knowledge---what the enterprise knows (system descriptions, terminology, data models, architectural decisions). These are complementary: ACE could optimize an agent's reasoning strategy while DDC provides the domain knowledge that strategy operates on.

  \item \textbf{Human role.} ACE is fully automated. DDC requires a human domain expert to provide answers that exist only as tribal knowledge. No amount of automated reflection can discover that ``the payment reconciliation service splits transactions across two ledgers because a regulatory change in 2019 required separate audit trails for domestic and cross-border payments''---this knowledge exists only in human heads.

  \item \textbf{Output structure.} ACE produces ``playbook bullets''---unstructured text appended to context. DDC produces typed, version-controlled knowledge entities with YAML frontmatter, explicit relationships, and a defined meta-model. This structure enables navigation, validation, and reuse across agents and humans.
\end{enumerate}

\subsection{Context Engineering}

The term ``context engineering'' was popularized by Karpathy~\citep{karpathy2025context} and further developed in Anthropic's guide to building effective agents~\citep{anthropic2025agents}, describing the practice of systematically determining what information an LLM agent needs in its context window. DDC can be understood as a \emph{methodology for context engineering}---it provides a systematic process for determining what context should exist, not just how to retrieve or arrange existing context.

\citet{karidi2025kgarchives} describe a demand-driven curation feedback loop in digital archives, where user search frequency drives metadata enrichment priorities. While applied to archival science rather than software agents, this work independently validates the core DDC principle: let demand (actual usage patterns) drive curation effort rather than attempting comprehensive coverage upfront.

\subsection{Novelty of DDC}

No existing work combines all five of DDC's core components: (a)~agent failure as the primary signal for knowledge gaps, (b)~human-curated structured knowledge entities, (c)~a typed entity meta-model for organizing domain knowledge, (d)~a convergence hypothesis for knowledge base completeness, and (e)~a formalized methodology with defined artifacts, analogous to TDD. Table~\ref{tab:novelty} summarizes the gap.

\begin{table}[ht]
\centering
\caption{Novelty gap analysis. No existing work combines all five DDC components.}
\label{tab:novelty}
\begin{tabular}{@{}lll@{}}
\toprule
\textbf{Component} & \textbf{Description} & \textbf{Found in Literature?} \\
\midrule
(a) & Agent failure as signal & Yes --- ACE, Reflexion, ExpeL \\
(b) & Human-curated structured knowledge & Partial --- AITL \\
(c) & Typed entity meta-model & No \\
(d) & Convergence hypothesis & No \\
(e) & Formalized named methodology & Partial --- ACE \\
\midrule
(a+b) & Failure-driven + human-curated KB & \textbf{No} \\
(a+b+c) & Above + typed meta-model & \textbf{No} \\
(a+b+c+d+e) & Full DDC synthesis & \textbf{No --- DDC is novel} \\
\bottomrule
\end{tabular}
\end{table}

\section{The DDC Methodology}
\label{sec:methodology}

\subsection{Overview}

DDC is a cyclic methodology for building enterprise knowledge bases. Each cycle is triggered by a real problem---a vendor integration question, an architectural design task, a cross-team coordination challenge. The agent attempts the problem with its current knowledge, fails, identifies what is missing, and the human curates only what is needed.

\begin{algorithm}[t]
\caption{DDC Cycle Protocol}\label{alg:ddc}
\begin{algorithmic}[1]
\Require Problem $P$, Knowledge Base $\mathit{KB}$ (initially $\emptyset$), Meta-Model $\mathcal{M}$
\Ensure Updated $\mathit{KB}$, Cycle Log $L$
\State Create sandbox workspace for $P$
\State $R_0 \gets \textsc{AgentAttempt}(P, \mathit{KB})$ \Comment{Agent tries with current context}
\State $C \gets \textsc{IdentifyGaps}(R_0)$ \Comment{Agent generates information checklist}
\State $A \gets \textsc{HumanFillGaps}(C)$ \Comment{Domain expert provides answers}
\State $E \gets \textsc{DraftEntities}(A, \mathcal{M})$ \Comment{Structure answers as typed entities}
\State $R_n \gets \textsc{AgentAttempt}(P, \mathit{KB} \cup E)$ \Comment{Agent re-attempts with new context}
\State $v \gets \textsc{HumanValidate}(R_n)$
\While{$v = \textit{rejected}$}
    \State $A' \gets \textsc{HumanCorrect}(R_n)$ \Comment{Expert provides corrections}
    \State Update $E$ from $A'$
    \State $R_n \gets \textsc{AgentAttempt}(P, \mathit{KB} \cup E)$
    \State $v \gets \textsc{HumanValidate}(R_n)$
\EndWhile
\State $\mathit{KB} \gets \mathit{KB} \cup E$ \Comment{Graduate validated entities}
\State $L \gets \textsc{LogCycle}(P, R_0, C, A, E, R_n)$
\State \Return $\mathit{KB}, L$
\end{algorithmic}
\end{algorithm}

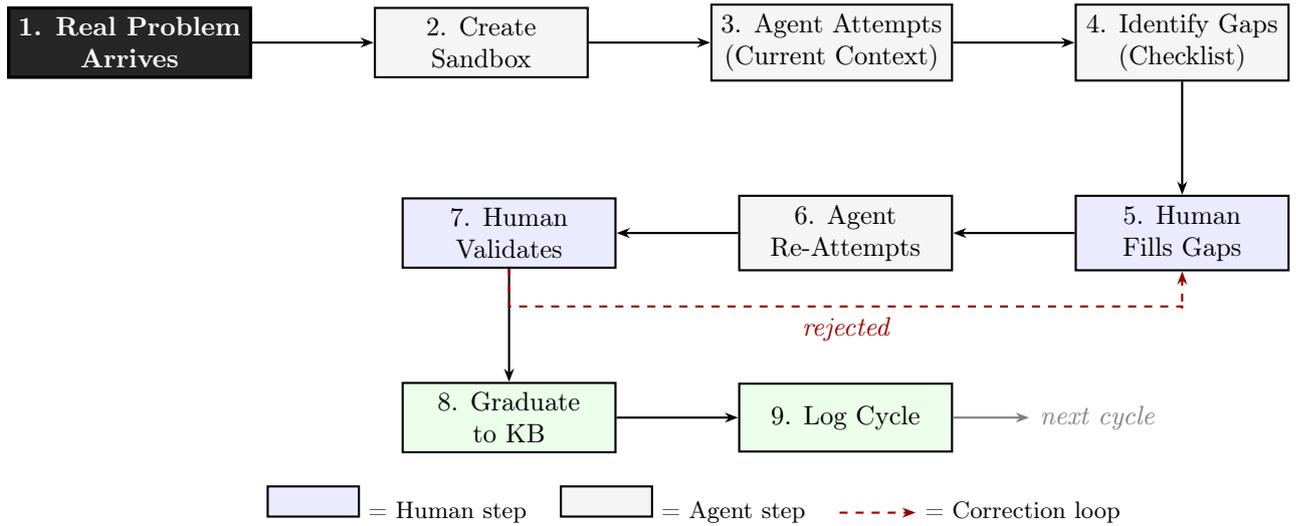
\begin{figure*}[t]
\centering
\begin{tikzpicture}[
  node distance=1.2cm and 1.6cm,
  stepbox/.style={rectangle, draw=black, thick, fill=gray!8, minimum width=2.8cm, minimum height=0.9cm, align=center, font=\small},
  humanbox/.style={rectangle, draw=black, thick, fill=blue!8, minimum width=2.8cm, minimum height=0.9cm, align=center, font=\small},
  startbox/.style={rectangle, draw=black, very thick, fill=black!85, text=white, minimum width=2.8cm, minimum height=0.9cm, align=center, font=\small\bfseries},
  endbox/.style={rectangle, draw=black, thick, fill=green!8, minimum width=2.8cm, minimum height=0.9cm, align=center, font=\small},
  arrow/.style={-{Stealth[length=5pt]}, thick},
  dasharrow/.style={-{Stealth[length=5pt]}, thick, dashed, red!60!black},
]
\node[startbox] (s1) {1. Real Problem\\Arrives};
\node[stepbox, right=of s1] (s2) {2. Create\\Sandbox};
\node[stepbox, right=of s2] (s3) {3. Agent Attempts\\(Current Context)};
\node[stepbox, right=of s3] (s4) {4. Identify Gaps\\(Checklist)};

\node[humanbox, below=1.5cm of s4] (s5) {5. Human\\Fills Gaps};
\node[stepbox, left=of s5] (s6) {6. Agent\\Re-Attempts};
\node[humanbox, left=of s6] (s7) {7. Human\\Validates};

\node[endbox, below=1.5cm of s7] (s8) {8. Graduate\\to KB};
\node[endbox, right=of s8] (s9) {9. Log Cycle};

\draw[arrow] (s1) -- (s2);
\draw[arrow] (s2) -- (s3);
\draw[arrow] (s3) -- (s4);
\draw[arrow] (s4) -- (s5);
\draw[arrow] (s5) -- (s6);
\draw[arrow] (s6) -- (s7);
\draw[arrow] (s7) -- (s8);
\draw[arrow] (s8) -- (s9);

\draw[dasharrow] (s7.south) -- ++(0,-0.5) -| node[pos=0.25, below, font=\small\itshape, red!60!black] {rejected} ($(s5.south)+(0,-0.5)$) -- (s5.south);

\draw[arrow, gray] (s9.east) -- ++(1,0) node[right, font=\small\itshape, gray] {next cycle};

\node[below=0.3cm of s9, xshift=-2cm, font=\footnotesize] {
  \tikz{\node[humanbox, minimum width=1.2cm, minimum height=0.4cm, font=\footnotesize] {}} = Human step \quad
  \tikz{\node[stepbox, minimum width=1.2cm, minimum height=0.4cm, font=\footnotesize] {}} = Agent step \quad
  \tikz{\draw[dasharrow] (0,0) -- (1,0);} = Correction loop
};
\end{tikzpicture}
\caption{The DDC cycle. A real problem triggers the loop. The agent attempts the problem, identifies knowledge gaps via an information checklist, and a human expert provides targeted answers. The agent re-attempts with new context. If the human rejects the output (dashed arrow), the correction loop repeats. Validated knowledge is graduated to the permanent knowledge base.}
\label{fig:ddc-loop}
\end{figure*}

\subsection{The DDC Cycle}

A single DDC cycle proceeds through nine steps:

\textbf{Step 1: Real Problem Arrives.} A concrete problem requiring domain expertise enters the system. This is not a synthetic exercise---it is a real question that needs answering for business operations to proceed. Examples include: a vendor asking integration questions, a design review requiring architectural reasoning, or a data model review requiring understanding of business rules.

\textbf{Step 2: Create Sandbox.} A workspace is created for the problem, containing the problem description and source documents. This isolation prevents work-in-progress from contaminating the curated knowledge base.

\textbf{Step 3: Agent Attempts (Zero/Current Context).} The agent attempts to solve the problem using only its general capabilities and whatever domain knowledge has been curated in previous cycles. In the first cycle, this means zero domain context.

\textbf{Step 4: Agent Identifies Gaps.} The agent produces an \emph{information checklist}---a structured list of what it needs to answer the problem. This checklist is organized by knowledge type: terminology needing definition, systems needing documentation, reference data needed, business logic needing explanation.

\textbf{Step 5: Human Fills Gaps.} A domain expert provides targeted answers for each checklist item. This is the critical human-in-the-loop step. The expert provides only what is needed---not a comprehensive brain dump, but minimum viable context for the specific problem.

\textbf{Step 6: Agent Re-Attempts.} With the new context, the agent attempts the problem again. If the output is still incorrect or incomplete, the cycle returns to Step~4.

\textbf{Step 7: Human Validates Output.} The domain expert reviews the agent's output for correctness. Minor corrections are incorporated. This validation step ensures that curated knowledge produces correct reasoning.

\textbf{Step 8: Graduate Content.} Validated knowledge is structured as typed entities (see Section~\ref{sec:metamodel}) and moved from the sandbox to the permanent knowledge base in the correct location based on entity type.

\textbf{Step 9: Log Cycle.} The cycle is recorded as a structured log entry containing: problem input, agent's initial performance, information checklist, entities created/updated, agent's final performance, and human review notes. These logs provide the data for convergence analysis.

\subsection{The Entity Meta-Model}
\label{sec:metamodel}

DDC knowledge is organized as typed entities stored as markdown files with YAML frontmatter. The meta-model defines the entity types shown in Table~\ref{tab:entity-types}.

\begin{table}[ht]
\centering
\caption{DDC entity types. Each entity is a markdown file with structured YAML frontmatter.}
\label{tab:entity-types}
\begin{tabular}{@{}llL{4.5cm}@{}}
\toprule
\textbf{Type} & \textbf{Description} & \textbf{Example} \\
\midrule
\texttt{jargon-business} & Business terminology & ``Fulfillment Unit'', ``Backordered'' \\
\texttt{jargon-tech} & Technical terminology & ``Dead Letter Queue'', ``Work Order Release'' \\
\texttt{system} & Software systems & ``Service Order Manager'', ``Picking Service'' \\
\texttt{capability} & Business capabilities & ``Service Fulfillment'', ``Order Capture'' \\
\texttt{data-model} & Core data structures & ``Parcel Shipping Data'' \\
\texttt{api} & API contracts & ``Picking-to-Routing Parcel API'' \\
\texttt{team} & Organizational units & ``Fulfillment Operations'' \\
\texttt{persona} & User roles and needs & ``Store Sales Staff'' \\
\texttt{decision} & Architectural decisions & ``Sync vs.\ Async Order Handoff'' \\
\texttt{process} & Business processes & ``Order-to-Delivery Flow'' \\
\bottomrule
\end{tabular}
\end{table}

\begin{figure}[t]
\centering
\begin{tikzpicture}[
  entity/.style={rectangle, draw=black, thick, fill=gray!8, minimum width=2.2cm, minimum height=0.7cm, align=center, font=\footnotesize\ttfamily},
  coreentity/.style={rectangle, draw=black, very thick, fill=black!8, minimum width=2.2cm, minimum height=0.7cm, align=center, font=\footnotesize\ttfamily\bfseries},
  rel/.style={-{Stealth[length=4pt]}, thick, gray!70!black},
  rellabel/.style={font=\tiny, gray!60!black, fill=white, inner sep=1pt},
  node distance=1.0cm and 1.8cm,
]
\node[coreentity] (sys) {system};
\node[coreentity, right=2.2cm of sys] (cap) {capability};
\node[entity, below=1.2cm of sys] (proc) {process};
\node[entity, above=1.2cm of sys] (api) {api};

\node[entity, left=2.2cm of sys] (jb) {jargon-biz};
\node[entity, above=1.2cm of jb] (jt) {jargon-tech};
\node[entity, below=1.2cm of jb] (dm) {data-model};

\node[entity, right=2.2cm of cap] (team) {team};
\node[entity, above=1.2cm of team] (per) {persona};
\node[entity, below=1.2cm of team] (dec) {decision};

\draw[rel] (sys) -- node[rellabel, above] {implements} (cap);
\draw[rel] (api) -- node[rellabel, left] {exposed\_by} (sys);
\draw[rel] (proc) -- node[rellabel, left] {uses} (sys);
\draw[rel] (cap) -- node[rellabel, above] {owned\_by} (team);
\draw[rel] (per) -- node[rellabel, right] {belongs\_to} (team);
\draw[rel] (dec) -- node[rellabel, right] {affects} (cap);
\draw[rel] (dm) -- node[rellabel, below] {used\_by} (sys);
\draw[rel] (jt) -- node[rellabel, above] {describes} (sys);
\draw[rel] (jb) to[out=30,in=150] node[rellabel, above] {defines} (cap);

\draw[rel, looseness=4] (sys.north west) to[out=150,in=210] node[rellabel, left] {depends\_on} (sys.south west);
\end{tikzpicture}
\caption{The DDC entity meta-model. Entity types are connected through typed relationships defined in YAML frontmatter. Bold entities (\texttt{system}, \texttt{capability}) form the core around which other entities cluster. The self-referencing arrow on \texttt{system} represents inter-system dependencies.}
\label{fig:metamodel}
\end{figure}
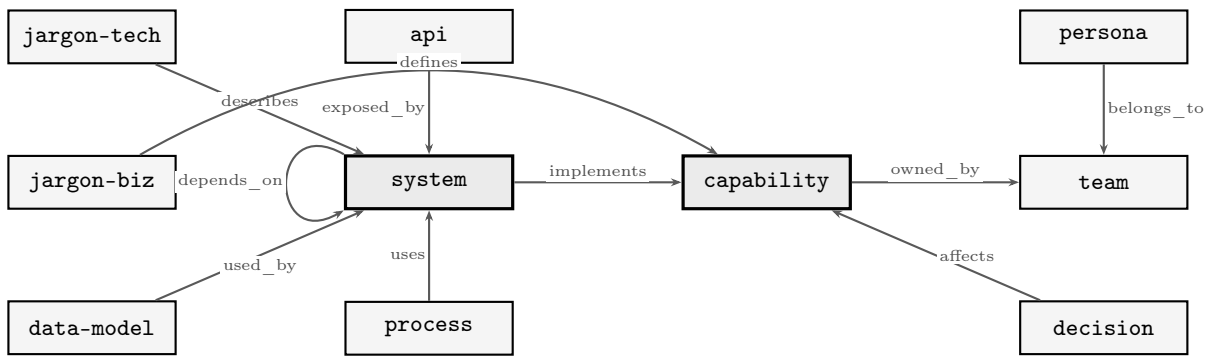

Figure~\ref{fig:metamodel} shows the relationships between entity types. Each entity file follows a consistent format:

\begin{lstlisting}[language=yaml,caption={Entity file format with YAML frontmatter and markdown body.}]
---
type: system
id: service-order-manager
name: Service Order Manager
description: Orchestrates service order lifecycle
status: active
related_systems: [provided-services-manager, message-broker]
implements_capability: service-fulfillment
---

# Service Order Manager

## Overview
[Prose description, purpose, ownership]

## Key Details
[Technical details, integration patterns, constraints]
\end{lstlisting}

This structure serves dual purposes: humans read the prose; machines parse the frontmatter for navigation, relationship traversal, and validation.

\subsection{The TDD Parallel}

Engineers will recognize DDC's structure in Test-Driven Development~\citep{beck2002tdd}. TDD writes a failing test, then the minimum code to pass. DDC gives the agent a failing problem, then the minimum context to succeed. Both methodologies share the principle of letting demand drive supply---you do not write all the tests upfront, and you do not curate all the context upfront.

This parallel extends to the development lifecycle:

\begin{itemize}
  \item \textbf{Red} (TDD) = Agent fails on problem (DDC)
  \item \textbf{Green} (TDD) = Agent succeeds with new context (DDC)
  \item \textbf{Refactor} (TDD) = Graduate content to proper knowledge base locations (DDC)
\end{itemize}

\subsection{The Convergence Hypothesis}
\label{sec:convergence}

We hypothesize that DDC exhibits convergence behavior: as more cycles are completed for a given domain role, each subsequent cycle requires fewer new entities because previous cycles have already curated overlapping knowledge.

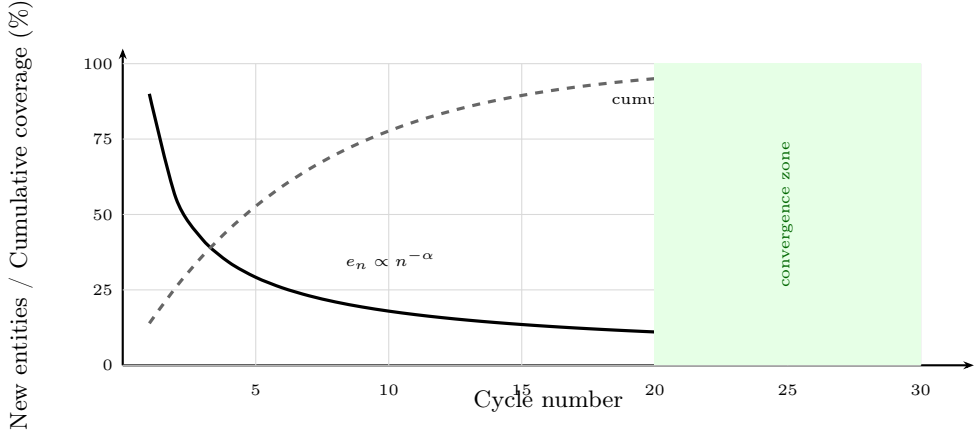
\begin{figure}[t]
\centering
\begin{tikzpicture}[xscale=0.35, yscale=0.04]
\draw[-{Stealth[length=4pt]}, thick] (0,0) -- (32,0);
\draw[-{Stealth[length=4pt]}, thick] (0,0) -- (0,105);

\node[font=\footnotesize] at (16,-12) {Cycle number};
\node[font=\footnotesize, rotate=90, anchor=south] at (-3,50) {New entities / Cumulative coverage (\%)};

\foreach \x in {5,10,15,20,25,30} {
  \node[font=\tiny, anchor=north] at (\x,-3) {\x};
  \draw[gray!30] (\x,0) -- (\x,100);
}
\foreach \y/\l in {0/0, 25/25, 50/50, 75/75, 100/100} {
  \node[font=\tiny, anchor=east] at (0,\y) {\l};
  \draw[gray!30] (0,\y) -- (30,\y);
}

\draw[black, very thick] plot[smooth, domain=1:30, samples=30]
  (\x, {90/(\x^0.7)});
\node[font=\tiny, anchor=west] at (8,35) {$e_n \propto n^{-\alpha}$};

\draw[black!60, very thick, dashed] plot[smooth, domain=1:30, samples=30]
  (\x, {100*(1 - exp(-0.15*\x))});
\node[font=\tiny, anchor=west] at (18,88) {cumulative coverage};

\fill[green!10] (20,0) rectangle (30,100);
\node[font=\tiny, green!40!black, rotate=90] at (25,50) {convergence zone};

\end{tikzpicture}
\caption{Hypothesized convergence curve. New entities per cycle (solid) follow a power-law decay; cumulative coverage (dashed) approaches an asymptote. The shaded region marks the hypothesized convergence zone (20--30 cycles) where most new problems can be solved with existing knowledge.}
\label{fig:convergence}
\end{figure}

Formally, if $e_n$ is the number of new entities created in cycle $n$, and $r_n$ is the number of entities reused from previous cycles, we expect:

\begin{itemize}
  \item $e_n$ follows a power-law decay: $e_n \propto n^{-\alpha}$ for some $\alpha > 0$
  \item $r_n / (e_n + r_n)$ increases monotonically---each cycle reuses a larger fraction of existing knowledge
  \item After approximately 20--30 cycles, $e_n \approx 0$ for most problems within the domain role's scope
\end{itemize}

This convergence is a hypothesis, not yet validated at scale. Section~\ref{sec:example} presents preliminary evidence from nine cycles in a retail fulfillment domain. Full validation with 20+ cycles is ongoing work.

\section{Worked Example: Retail Order Fulfillment}
\label{sec:example}

To demonstrate DDC in practice, we present a worked example in the retail e-commerce order fulfillment domain, targeting an SRE (Site Reliability Engineering) incident management agent. The domain is synthetic but representative---entities, systems, and incident patterns reflect common patterns in large-scale e-commerce fulfillment operations but do not reflect any specific organization.

\subsection{Domain Setup}

The order fulfillment domain involves processing customer orders from placement through delivery. Key activities include order capture, inventory allocation, warehouse picking, routing to delivery carriers, and last-mile delivery coordination. The domain spans multiple interconnected systems, involves both online and in-store channels, and operates across multiple geographic markets---characteristics shared with most enterprise operations.

The target agent role is an \emph{SRE incident management agent}: given a production incident report (orders stuck, deliveries failing, systems unresponsive), the agent must diagnose root causes, identify affected systems, and recommend remediation. This requires deep knowledge of system interdependencies, message flows, and failure modes---exactly the kind of knowledge that exists as tribal knowledge among experienced on-call engineers.

We start with an empty knowledge base and run nine DDC cycles triggered by realistic incident scenarios.

\subsection{Cycle 1: Service Order Queue Contention}

\textbf{Problem:} Thousands of service orders are stuck in ``Ready to Assign'' status for over a day. The fulfillment platform is not processing them.

\textbf{Agent Before (Zero Context):} The agent produced generic troubleshooting advice: check queue consumers, verify message broker health, look at logs. It could not identify which specific systems are involved in service order processing, what ``Ready to Assign'' means in this domain, or what the expected message flow looks like.

\textbf{Information Checklist Generated:}
\begin{itemize}[nosep]
  \item What systems participate in service order fulfillment?
  \item What does ``Ready to Assign'' status mean and what triggers it?
  \item What is the message flow from order creation to fulfillment assignment?
  \item What message broker infrastructure connects these systems?
  \item What is a ``service booking'' and how does it relate to orders?
\end{itemize}

\textbf{Human Input:} The domain expert explained that service orders flow through the Service Order Manager to the Provided Services Manager, which assigns them to an External Routing Provider via a message broker. ``Ready to Assign'' is an intermediate state indicating the order is validated but not yet dispatched. The expert described the specific queue architecture and consumer group configuration.

\textbf{Entities Created:} 8---service-order-manager, provided-services-manager, external-routing-provider, message-broker (\texttt{system}/\texttt{platform}); service-fulfillment-flow (\texttt{process}); service-booking, ready-to-assign (\texttt{jargon-business}); service-fulfillment (\texttt{capability}).

\textbf{Agent After:} The agent correctly identified the queue contention point between specific systems, explained the expected message flow, and recommended targeted diagnostics for the consumer group configuration. The response contained system names, message flow stages, and specific failure modes rather than generic advice.

\subsection{Cycle 6: Cross-Region Deployment Error (Rejected Attempts)}

We highlight cycle~6 to illustrate how DDC handles agent failures that persist even after initial curation---and why logging rejected attempts is critical.

\textbf{Problem:} Customers in a European market see all orders marked as ``backordered'' despite inventory being available in warehouses.

\textbf{Agent After---Attempt 1 (Rejected):} The agent hypothesized an inventory synchronization delay at the application logic layer. The human rejected this: the root cause was not logic but \emph{data}---a deployment script intended for a different regional environment was accidentally executed against the European environment, corrupting configuration data.

\textbf{Agent After---Attempt 2 (Rejected):} The agent fabricated a plausible-sounding root cause involving a ``data replication pipeline'' that does not exist in this domain. It also confused warehouse identifiers with technology environment names. The human corrected both errors.

\textbf{Agent After---Attempt 3 (Accepted):} With corrected context, the agent accurately described the root cause: an XML deployment script run against the wrong regional compartment corrupted the Service Order Manager's configuration, causing all inventory checks to return false. The agent also identified the missing safeguard (a four-eyes review principle for cross-environment deployments) and recommended remediation.

\textbf{Entities Created:} 7 new entities including backordered, available-to-promise (\texttt{jargon-business}); compartment-environment, cross-compartment-deployment-error, four-eyes-principle (\texttt{jargon-tech}); inventory-allocation (\texttt{capability}); distribution-point (\texttt{jargon-business}). Two entities the agent fabricated were explicitly deleted.

This cycle demonstrates a key DDC property: \emph{the correction loop is the most valuable data}. The agent's fabrication in attempt~2 revealed that it will confidently generate plausible but false domain knowledge when uncertain---precisely the failure mode that motivates human curation.

\subsection{Cycles 2--9: Accumulating Knowledge}

Table~\ref{tab:cycles} summarizes all nine cycles.

\begin{table}[ht]
\centering
\caption{DDC cycle summary for the retail fulfillment domain. New entities decrease and reused entities increase across cycles. Cycles marked with~$\dagger$ had rejected agent attempts requiring correction loops.}
\label{tab:cycles}
\begin{tabular}{@{}clcccc@{}}
\toprule
\textbf{Cycle} & \textbf{Incident} & \textbf{New} & \textbf{Updated} & \textbf{Reused} & \textbf{Time (min)} \\
\midrule
001 & Service order queue contention & 8 & 0 & 0 & 30 \\
002 & Orders not reaching carrier & 4 & 0 & 3 & 25 \\
003 & Package weight mispricing & 4 & 3 & 5 & 30 \\
004$^\dagger$ & Orders missing after handoff & 5 & 4 & 8 & 30 \\
005$^\dagger$ & In-store orders failing & 6 & 3 & 7 & 30 \\
006$^\dagger$ & Cross-region deployment error & 7 & 4 & 6 & 45 \\
007 & Checkout delivery page outage & 8 & 0 & 3 & 30 \\
008$^\dagger$ & Warehouse pick instructions missing & 2 & 2 & 6 & 25 \\
009 & Regional booking outage & 2 & 3 & 6 & 25 \\
\bottomrule
\end{tabular}
\end{table}

\begin{figure}[t]
\centering
\begin{tikzpicture}[
  xscale=0.85, yscale=0.45,
]
\draw[gray!20] (0,0) grid[xstep=1,ystep=1] (9.5,9);

\draw[-{Stealth[length=4pt]}, thick] (0,0) -- (0,9.5);
\node[font=\footnotesize, rotate=90, anchor=south] at (-0.8,4.5) {New entities created};
\foreach \y in {0,2,4,6,8} {
  \node[font=\tiny, anchor=east] at (0,\y) {\y};
}

\draw[-{Stealth[length=4pt]}, thick] (9.5,0) -- (9.5,9.5);
\node[font=\footnotesize, rotate=-90, anchor=south] at (10.3,4.5) {Reused entities};
\foreach \y in {0,2,4,6,8} {
  \node[font=\tiny, anchor=west] at (9.5,\y) {\y};
}

\draw[-{Stealth[length=4pt]}, thick] (0,0) -- (10,0);
\node[font=\footnotesize] at (4.75,-1.2) {DDC Cycle};
\foreach \x/\n in {0.5/1, 1.5/2, 2.5/3, 3.5/4, 4.5/5, 5.5/6, 6.5/7, 7.5/8, 8.5/9} {
  \node[font=\tiny] at (\x,-0.5) {\n};
}

\foreach \x/\h in {0.5/8, 1.5/4, 2.5/4, 3.5/5, 4.5/6, 5.5/7, 6.5/8, 7.5/2, 8.5/2} {
  \fill[black!50] (\x-0.3,0) rectangle (\x+0.3,\h);
}

\draw[black, very thick, mark=*] plot coordinates {(0.5,0) (1.5,3) (2.5,5) (3.5,8) (4.5,7) (5.5,6) (6.5,3) (7.5,6) (8.5,6)};
\foreach \x/\y in {0.5/0, 1.5/3, 2.5/5, 3.5/8, 4.5/7, 5.5/6, 6.5/3, 7.5/6, 8.5/6} {
  \fill[black] (\x,\y) circle (3pt);
}

\draw[gray, thin, dashed] (6.5,8) -- (6.5,9);
\node[font=\tiny, gray, anchor=south, align=center] at (6.5,9) {new\\subsystem};

\fill[black!50] (1,8.5) rectangle (1.6,8.8);
\node[font=\tiny, anchor=west] at (1.7,8.65) {New entities};
\draw[black, very thick] (4,8.65) -- (4.6,8.65);
\fill[black] (4.3,8.65) circle (2.5pt);
\node[font=\tiny, anchor=west] at (4.7,8.65) {Reused entities};

\end{tikzpicture}
\caption{Entity creation and reuse across nine DDC cycles. Bars show new entities created (left axis); the line shows entities reused from previous cycles (right axis). New entities generally decrease while reuse increases, consistent with the convergence hypothesis. Cycle~7 is an outlier---it introduced a previously uncovered subsystem.}
\label{fig:fulfillment-convergence}
\end{figure}
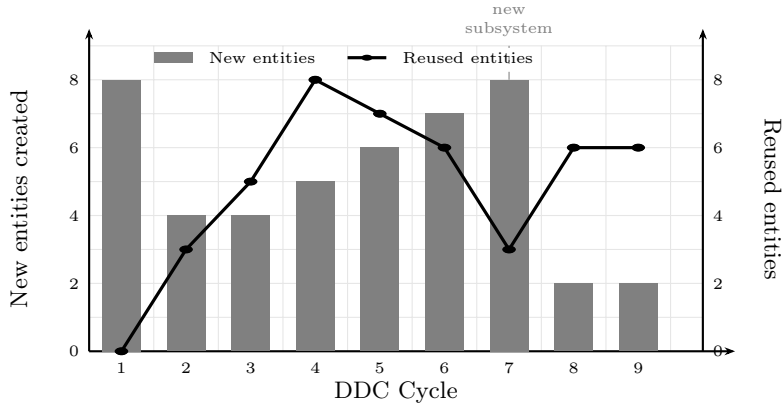

Three patterns emerge from these nine cycles:

\begin{enumerate}
  \item \textbf{New entities per cycle generally decrease:} From 8 in cycle~1 to 2 in cycles~8--9. The agent requires less new knowledge as foundational entities accumulate. Cycle~7 is an outlier (8 new entities) because it introduced a previously uncovered subsystem (checkout and delivery options orchestration).

  \item \textbf{Reused entities per cycle increase:} From 0 in cycle~1 to 6 in cycles~8--9. By the later cycles, the agent leverages existing knowledge of systems, message flows, and failure patterns to diagnose new incidents.

  \item \textbf{Correction loops decrease:} Four of the first six cycles required rejected attempts and correction loops. Of the last three cycles, none required corrections---the accumulated knowledge base provided sufficient context for accurate first-attempt diagnoses.
\end{enumerate}

The reuse ratio $r_n / (e_n + r_n)$ increases from 0.0 in cycle~1 to 0.75 in cycle~9. Total curation time across nine cycles was 270 minutes (4.5 hours), averaging 30 minutes per cycle.

\subsection{Knowledge Base Structure}

After nine cycles, the knowledge base contains 46 entities organized across the DDC meta-model, as shown in Table~\ref{tab:entity-breakdown}.

\begin{table}[ht]
\centering
\caption{Entity breakdown by type after nine DDC cycles.}
\label{tab:entity-breakdown}
\begin{tabular}{@{}lcL{5cm}@{}}
\toprule
\textbf{Entity Type} & \textbf{Count} & \textbf{Examples} \\
\midrule
\texttt{system} & 12 & service-order-manager, warehouse-management-system, picking-service, delivery-options-orchestrator \\
\texttt{jargon-tech} & 12 & dead-letter-queue, cross-compartment-deployment-error, work-order-release, no-autoscaling-pattern \\
\texttt{jargon-business} & 9 & ready-to-assign, fulfillment-unit, backordered, available-to-promise, weight-based-pricing \\
\texttt{process} & 5 & order-to-delivery-flow, service-fulfillment-flow, checkout-delivery-options-flow \\
\texttt{capability} & 4 & service-fulfillment, order-capture, delivery-arrangement, inventory-allocation \\
\texttt{platform} & 1 & message-broker \\
\texttt{persona} & 1 & store-sales-staff \\
\texttt{data-model} & 1 & parcel-shipping-data \\
\texttt{api} & 1 & picking-to-routing-parcel-api \\
\midrule
\textbf{Total} & \textbf{46} & \\
\bottomrule
\end{tabular}
\end{table}

Notable characteristics of the resulting knowledge base: (a)~\texttt{system} and \texttt{jargon-tech} entities dominate, reflecting the SRE focus---incident management requires understanding system boundaries and technical failure patterns; (b)~\texttt{process} entities capture the message flows that are critical for tracing where orders get stuck; (c)~the entity type distribution would likely differ for a different domain role (e.g., a product owner agent would produce more \texttt{capability} and \texttt{jargon-business} entities).

The full knowledge base, including all entity files, cycle logs, and meta-model definitions, is available as a reproducible artifact.\footnote{\url{https://github.com/ea-toolkit/ddc}}

\section{Scaling DDC: A Proposed Architecture}
\label{sec:scaling}

The DDC methodology as described in Section~\ref{sec:methodology} requires a human domain expert to manually curate every entity. While effective for proving the concept, this creates a bottleneck in enterprise environments where multiple teams encounter problems simultaneously and domain experts' time is scarce.

We propose a scaling architecture that introduces semi-automated curation while preserving human governance as a quality gate.

\subsection{The Manual Curation Bottleneck}

In the base DDC methodology, the human performs three roles: (1)~\emph{Information Provider}---answering the agent's checklist questions; (2)~\emph{Entity Author}---structuring answers as typed entities with frontmatter; and (3)~\emph{Validator}---reviewing the agent's output for correctness.

Roles~2 and~3 can be partially automated. The agent can draft entity files from human-provided answers (structured according to the meta-model), and validation can be decomposed into automated checks (schema validation, relationship consistency) and human judgment (domain correctness).

\subsection{Semi-Automated Curation}

In the proposed architecture, the DDC cycle is augmented:

\begin{itemize}
  \item \textbf{Steps 1--5} remain unchanged---real problems drive the cycle, agents identify gaps, humans provide domain answers.
  \item \textbf{Step 6 (Enhanced):} The agent drafts structured entities from the human's answers, following the meta-model templates. The human reviews and corrects rather than authors from scratch.
  \item \textbf{Step 8 (Enhanced):} Graduated content is submitted as a pull request to the centralized knowledge base. Automated checks validate schema, relationships, and naming conventions. A human domain expert reviews for correctness and approves.
\end{itemize}

\subsection{The 20/80 Curation Strategy}

Not all domain knowledge requires deep curation. We propose a 20/80 strategy:

\begin{itemize}
  \item \textbf{Deep curation (20\%):} Entities that the agent needs to reason about---systems, business logic, architectural decisions, key terminology. These get full prose descriptions, explicit relationships, and detailed metadata.
  \item \textbf{Lightweight stubs (80\%):} Entities that exist for reference but are not central to reasoning---peripheral systems, rarely-used terminology, edge-case reference data. These get a one-line description and links to external sources.
\end{itemize}

The problem determines what falls in the 20\%. Whatever the agent needs to answer the immediate question gets deep curation. Everything else gets a stub. Over time, stubs that are repeatedly referenced across cycles get promoted to deep curation---another form of demand-driven prioritization.

\subsection{Parallel Problem-Solving with Centralized Governance}

In an enterprise with multiple teams, DDC cycles can run in parallel:

\begin{itemize}[nosep]
  \item Team~A encounters an order fulfillment question $\rightarrow$ triggers cycle~A
  \item Team~B encounters a delivery routing question $\rightarrow$ triggers cycle~B
  \item Both cycles produce curated entities submitted as pull requests to the same centralized knowledge base
  \item A domain steward reviews PRs for consistency, deduplication, and correctness
  \item Merge conflicts (two teams curating the same entity differently) are resolved through the PR review process
\end{itemize}

This architecture mirrors how software teams manage shared codebases through version control and code review. The knowledge base is treated as a shared codebase with the same governance---branching, pull requests, review, and merge.

\subsection{Limitations of the Proposed Architecture}

This scaling architecture is proposed but not yet validated. Key open questions include: Does human review become the new bottleneck at scale? Can centralized governance maintain entity quality across dozens of parallel contributors? Is the 20/80 ratio stable across domains or domain-dependent? Validating this architecture is planned future work.

\section{Discussion}
\label{sec:discussion}

\subsection{What DDC Claims vs.\ What Needs Validation}

DDC makes claims at three levels of evidence:

\begin{enumerate}
  \item \textbf{Demonstrated:} The DDC methodology (Section~\ref{sec:methodology}) has been applied in both synthetic (Section~\ref{sec:example}) and real enterprise settings. The cycle mechanics work: agents do identify knowledge gaps, human-curated context does improve agent performance, and entities are reused across cycles.

  \item \textbf{Preliminary evidence:} The convergence trend (decreasing new entities, increasing reuse) is observed in nine cycles. This is suggestive but not statistically significant. Validation with 20+ cycles is ongoing.

  \item \textbf{Proposed:} The scaling architecture (Section~\ref{sec:scaling}) is a design, not an implementation. Its effectiveness is hypothetical.
\end{enumerate}

\subsection{Relationship to Reinforcement Learning}

DDC can be understood through a reinforcement learning lens. The agent's attempt is a policy execution. The human's validation is a reward signal. The knowledge base update is a policy update---not through weight modification but through context modification. The convergence hypothesis parallels the decreasing loss in RL training. This framing is conceptual rather than formal, but it suggests connections to the reward shaping and curriculum learning literature that may prove productive.

\subsection{Limitations}

\textbf{Human dependency.} DDC requires access to domain experts willing to answer agent generated checklists. In organizations where expert time is scarce, this may limit adoption. The semi-automated architecture (Section~\ref{sec:scaling}) partially addresses this by reducing human effort from authoring to reviewing.

\textbf{Single-domain validation.} The retail fulfillment example, while representative, is synthetic. The convergence hypothesis requires validation across multiple real enterprise domains.

\textbf{Subjectivity of curation.} Different humans may curate different entities for the same problem. DDC does not prescribe how to resolve disagreements in curation, beyond the PR review process proposed in Section~\ref{sec:scaling}.

\textbf{No formal convergence proof.} The convergence hypothesis is empirically motivated but lacks formal analysis. Establishing theoretical bounds on convergence rates is future work.

\section{Conclusion and Future Work}
\label{sec:conclusion}

We have presented Demand-Driven Context (DDC), a methodology for building enterprise knowledge bases by using agent failure on real problems as the primary signal for what to curate. DDC combines failure-driven knowledge acquisition, human-curated structured entities, a convergence hypothesis, and a formalized development methodology analogous to TDD. Through a worked example in retail order fulfillment incident management, we demonstrated that nine DDC cycles produce a reusable knowledge base of 46 entities with increasing reuse and decreasing correction loops across cycles.

DDC contributes a missing piece to the context engineering landscape: while existing work focuses on \emph{retrieving} existing context (RAG) or \emph{optimizing} agent execution strategies (ACE, Reflexion), DDC addresses the upstream question of \emph{what context should exist in the first place}.

\subsection*{Future Work}

\textbf{Convergence validation.} Ongoing work is collecting DDC cycle data from a real enterprise domain to validate the convergence hypothesis with 20+ cycles. We aim to characterize the convergence curve and determine whether the 20--30 cycle threshold holds across domains.

\textbf{Scaling architecture implementation.} The semi-automated architecture with PR governance (Section~\ref{sec:scaling}) will be implemented and evaluated in a multi-team enterprise setting.

\textbf{Multi-agent knowledge sharing.} DDC currently targets one domain role at a time. Investigating how knowledge bases for different roles (architect, product owner, engineer) overlap and can be shared is an open question.

\textbf{Formal convergence analysis.} Establishing theoretical bounds on DDC convergence rates, potentially through connections to curriculum learning and active learning theory.

\textbf{Automated curation quality metrics.} Developing metrics to assess the quality of curated entities beyond human review---measuring coherence, completeness, and consistency programmatically.

\bibliographystyle{plainnat}
\bibliography{references}

\appendix

\section{Entity Meta-Model Specification}
\label{app:metamodel}

The complete entity meta-model, including all type definitions, relationship types, and file format specifications, is available in the accompanying repository under \texttt{meta/entity-types.yaml} and \texttt{meta/relationship-types.yaml}.

\section{DDC Cycle Log Format}
\label{app:cyclelog}

Each DDC cycle is logged using the following structured format:

\begin{lstlisting}[language=yaml,caption={DDC cycle log format.}]
---
cycle_id: "NNN"
problem_name: "Short problem description"
date_started: YYYY-MM-DD
date_completed: YYYY-MM-DD
time_spent_minutes: N
entities_created: N
entities_updated: N
domain: "domain-name"
---

# Cycle NNN: Problem Name

## Problem Input
## Agent Before (Zero/Current Context)
## Information Checklist
## Human Answers
## Entities Curated
## Agent After (With Context)
## Human Review
## Context Reuse Notes
\end{lstlisting}

\section{Example DDC Cycle Transcripts}
\label{app:transcripts}

We present excerpts from two DDC cycles to illustrate the methodology in practice: a clean first cycle (Cycle~1) and a cycle with multiple correction loops (Cycle~6).

\subsection*{Cycle 1: Service Order Queue Contention}

\textbf{Problem Input:}
\begin{quote}
\small
The service fulfillment platform processes customer-booked services (installation, delivery). After a routine deployment, thousands of service orders got stuck in ``Ready to Assign'' status, meaning customers had confirmed appointments but no service provider was dispatched. The issue went undetected for over a day. Analyze the root cause and propose architectural safeguards.
\end{quote}

\textbf{Agent Response (Zero Context):}
\begin{quote}
\small\itshape
``The likely root cause is some kind of message routing issue---perhaps a deployment changed a queue configuration, causing messages to be consumed by the wrong service.''

Confidence: 1/5. No knowledge of systems, message flow, order lifecycle, or business impact. Could not name any system or propose specific safeguards.
\end{quote}

\textbf{Information Checklist Generated (excerpt):}
\begin{enumerate}[nosep]
\small
  \item Service order management system --- \textit{type: system}
  \item Service provider management system --- \textit{type: system}
  \item External routing/delivery system --- \textit{type: system}
  \item Message broker platform and queue organization --- \textit{type: platform}
  \item ``Ready to Assign'' status meaning in order lifecycle --- \textit{type: jargon-business}
  \item Order-to-service-delivery process flow --- \textit{type: process}
\end{enumerate}

\textbf{Human Answers (excerpt):}
\begin{quote}
\small
Service orders flow through the Service Order Manager to the Provided Services Manager, which assigns them to an External Routing Provider via a message broker. ``Ready to Assign'' is an intermediate state indicating the order is validated but not yet dispatched to the routing provider. The queue architecture uses consumer groups on the message broker.
\end{quote}

\textbf{Entities Curated:} 8 entities created (4 systems, 1 process, 2 jargon-business, 1 capability).

\textbf{Agent Response (With Context):}
\begin{quote}
\small\itshape
Root cause: Competing consumer conflict on the message broker. A deployment of an unrelated service (AttachmentService) bound to the Provided Services Manager's queue, splitting messages between two consumers.

Detection gap: No order flow monitoring, no queue depth alerts, no state timeout alerts on ``Ready to Assign'' status.

Chronic pattern: The same symptom (orders stuck in Ready to Assign) has recurred 2+ years with different root causes, indicating a systemic design issue rather than isolated incidents.

Proposed safeguards: (1) Queue registry with deployment-time validation, (2) order flow monitoring comparing inflow vs.\ outflow rates, (3) state timeout alerts, (4) integration testing for queue bindings, (5) dead letter queues for unprocessable messages.

Confidence: 4/5.
\end{quote}

\bigskip
\hrule
\bigskip

\subsection*{Cycle 6: EU Orders Falsely Backordered (Correction Loop)}

\textbf{Problem Input:}
\begin{quote}
\small
Orders placed through the Service Order Manager are immediately getting backordered across the EU region, even when inventory is available. Customers see their orders stuck with no delivery date.
\end{quote}

\textbf{Agent Response (Current Context):}
\begin{quote}
\small\itshape
The knowledge base has no coverage of inventory/availability systems, backorder logic, or the concept of ``backordered'' as an order status. Guessed that ``something in the Service Order Manager or a system it checks is returning incorrect inventory/availability data for EU markets.''

Confidence: 1/5.
\end{quote}

\textbf{Agent After---Attempt 1 (Rejected):}
\begin{quote}
\small\itshape
Hypothesized that a ``market-to-DP mapping'' misconfiguration caused the availability-to-promise logic to check the wrong fulfillment locations. Claimed confidence: 3/5.
\end{quote}

\textbf{Human Correction:}
\begin{quote}
\small
The agent was looking at the wrong layer---the availability logic itself was fine. The agent built ``vocabulary without insight.'' Your confidence should still be 1.5/5.
\end{quote}

\textbf{Agent After---Attempt 2 (Rejected):}
\begin{quote}
\small\itshape
Treated the expert's diagnostic questions as confirmed root cause. Created a fabricated entity describing a data replication pipeline that does not exist. Wrote detailed incident narratives claiming data capture misconfiguration was the confirmed cause. Confused distribution point warehouse codes (physical identifiers like DP074) with a data replication technology. This was worse than Attempt~1---it polluted the knowledge base with fabricated facts presented as truth. Claimed confidence: 3/5.
\end{quote}

\textbf{Human Correction:}
\begin{quote}
\small
Actual root cause: A cutover team ran an XML deployment script intended for the CN (China) compartment against the EU compartment. This corrupted the Service Order Manager's inventory module configuration. Availability-to-promise ran correctly against corrupted data, producing false backorders. The fabricated entity is entirely wrong and must be deleted.
\end{quote}

\textbf{Agent After---Attempt 3 (Accepted):}
\begin{quote}
\small\itshape
Root cause: Cutover team ran XML deployment script intended for the CN compartment against the EU compartment, corrupting the Service Order Manager's inventory module configuration.

Mechanism: Availability-to-promise checked corrupted data, all checks returned false, triggering mass backorders.

Pattern: ``Correct logic, corrupted configuration''---every system works correctly in isolation; the failure is in the data/config layer.

Missing safeguards: No compartment validation on deployment scripts, no four-eyes principle for cross-environment deployments, no pre/post deployment health checks.

Confidence: 4/5.
\end{quote}

\medskip
\noindent\textit{Note:} The two fabricated entities from Attempt~2 were explicitly deleted. Seven new entities were curated from the corrected understanding. This cycle demonstrates why DDC requires human curation: the agent confidently generates plausible but false domain knowledge when uncertain, and only a domain expert can distinguish fabrication from fact.

\end{document}